\documentclass[sigconf]{acmart}

\renewcommand\footnotetextcopyrightpermission[1]{} 
\usepackage{times}
\usepackage{latexsym}
\usepackage{graphicx}
\usepackage{amsmath}
\usepackage{url}
\usepackage{enumitem}
\usepackage{tabu}
\usepackage{balance}
\usepackage[colorinlistoftodos,prependcaption,textsize=tiny]{todonotes}
\usepackage{tabularx, environ}
\usepackage{titlesec}
\usepackage{float}
\usepackage{xcolor}

\usepackage{algorithm}
\usepackage{algcompatible}
\usepackage[noend]{algpseudocode}
\makeatletter
\def\BState{\State\hskip-\ALG@thistlm}
\makeatother

\algdef{SE}[SUBALG]{Indent}{EndIndent}{}{\algorithmicend\ }%
\algtext*{Indent}
\algtext*{EndIndent}

\AtBeginDocument{%
  \providecommand\BibTeX{{%
    \normalfont B\kern-0.5em{\scshape i\kern-0.25em b}\kern-0.8em\TeX}}}

\setcopyright{acmcopyright}

\newcommand\METHOD{Co-Decomp~}
\newcommand\PHMDT{PHM~}
\newcommand\FLUDT{FLU~}
\newcommand\ADRDT{ADR~}
\newcommand\CRISISDT{CRISIS~}

\renewcommand{\textrightarrow}{$\rightarrow$}

\begin{document}

\copyrightyear{2020}
\acmYear{2020}
\acmConference[WWW '20]{Proceedings of The Web Conference 2020}{April 20--24, 2020}{Taipei, Taiwan}
\acmBooktitle{Proceedings of The Web Conference 2020 (WWW '20), April 20--24, 2020, Taipei, Taiwan}
\acmPrice{}
\acmDOI{10.1145/3366423.3380304}
\acmISBN{978-1-4503-7023-3/20/04}

\title{Domain-Guided Task Decomposition with Self-Training for Detecting Personal Events in Social Media}


\author{Payam Karisani}
\affiliation{%
  \institution{Emory University}
}
\email{payam.karisani@emory.edu}

\author{Joyce C. Ho}
\affiliation{%
  \institution{Emory University}
}
\email{joyce.c.ho@emory.edu}

\author{Eugene Agichtein}
\affiliation{%
  \institution{Emory University}
}
\email{eugene.agichtein@emory.edu}


\begin{abstract}
Mining social media content for tasks such as detecting personal experiences or events, suffer from lexical sparsity, insufficient training data, and inventive lexicons. To reduce the burden of creating extensive labeled data and improve classification performance, we propose to perform these tasks in two steps: 1. Decomposing the task into domain-specific sub-tasks by identifying key concepts, thus utilizing human domain understanding; and 2. Combining the results of learners for each key concept using co-training to reduce the requirements for labeled training data. We empirically show the effectiveness and generality of our approach, Co-Decomp, using three representative social media mining tasks, namely Personal Health Mention detection, Crisis Report detection, and Adverse Drug Reaction monitoring. The experiments show that our model is able to outperform the state-of-the-art text classification models--including those using the recently introduced BERT model--when small amounts of training data are available.
\end{abstract}




\maketitle

\section{Introduction} \label{sec:intro}

Social networks, such as Twitter and Facebook, have become inseparable parts of societies. A broad spectrum of topics are shared and discussed in the networks every day, and this has turned them into a suitable means for the online public monitoring. The applications include, but not limited to, consumer opinion mining \cite{opinion}, stock market prediction \cite{stock}, sarcasm detection \cite{sarcasm}, and user reputation management \cite{OLFinder}. These cases signify that social networks, e.g., Twitter, went beyond their initial purpose years ago--which was being simple personal messaging tools\footnote{\url{https://www.nytimes.com/2010/10/31/technology/31ev.html}}. Personal Event Detection is an example of the online public monitoring. For instance, in the case of Personal Health Mention detection \cite{healthbook}, the aim is to mine and track any individual health event. Scalability, real-time surveillance, and rapid response to potential outbreaks are the main advantages of this task when it is used inside a public health monitoring system. Another example is Crisis Report detection \cite{crisis-survey} through social media, which aims to mine user postings and alert humanitarian institutions and agencies during natural disasters. 




Even though social networks are a valuable source of information, mining user postings comes with several challenges. For instance, the tasks usually suffer from the lack of enough training data \cite{PHM}. Even in the cases that there is enough resources to construct a training set, the class distributions might be highly imbalanced \cite{adr,wellness}. Thus, having machine learning models to perform well in this data scarce environment is of great value. 


In classification tasks a common practice is to first extract a set of features, either manually or through representation learning, and then train a classifier over the resulting feature vectors. While training a single classifier over the entire content is a standard practice, an end-to-end classifier may require substantial amount of annotated data. Instead, for a subset of tasks, we can use domain knowledge to decompose the problem into a set of sub-tasks, and use a separate learner to tackle each one individually. This can lead to the development of models which are equipped with domain understanding and require less training data. For instance, if the task is cancer surveillance on the Twitter website, in the tweet \textit{``I Just went to my Oncology appointment at the Hospital!!! Praying that it's not cancer''}, we might be able to infer the class label from the contextual information of either the word ``I'' or ``cancer''. Therefore, we can solve each classification problem individually and aggregate the results.

We propose \METHOD\!, a semi-supervised model that can classify short text for problems with a set of sub-tasks. While our model can be potentially applied to any problem that is centered around a group of concepts or entities, we focus on three personal event detection tasks; because they usually suffer from the lack of training data and imbalanced class distributions, as mentioned earlier. Namely, we focus on Personal Health Mention detection \cite{PHM}, Crisis Report detection \cite{crisis-survey}, and Adverse Drug Reaction monitoring \cite{adr}, and show that \METHOD can outperform state-of-the-art classifiers in semi-supervised settings. In summary, our contributions are:
\begin{itemize}
    \item We propose Key Concept Sets to decompose a particular category of text classification problems, referred to as decomposable problems, into a set of sub-tasks.
    \item We introduce a co-training model to effectively utilize the problem decomposition, and reduce the need for training data.
    \item We show that a category of personal event detection tasks fall into the class of decomposable problems. We carry out comprehensive experiments on four datasets, and show that our model reduces the need for training data, and can outperform state-of-the-art classifiers in the low data regime.
\end{itemize}
Together, these contributions significantly advance the state of the art in the personal event detection and related tasks. Next, we review the related work to place our contributions in context.


\section{Related Work} \label{sec:rel-work}

Our model falls into the category of divide-and-conquer algorithms, and this family of algorithms have been employed in text classification before. For example, a pipeline of filtering steps have been applied to documents in order to filter out the confidently negative ones \cite{wellness}. The main difference between our model and the pipelining approach is that we initially decompose the task into a set of sub-tasks that can be complementary, whereas in the case of pipelining, the final classifier still needs to tackle the same initial task. Additionally, our decomposition reduces the need for training data such that the task can be solved in semi-supervised settings. Our model is also deeply connected to the information extraction \cite{distantsup}, relation classification \cite{att-rel-class}, and semantic role labeling \cite{srl} tasks in natural language processing. In addition to be agnostic towards the number of entities and their relation type, which are pivotal in the mentioned tasks, our proposal is mainly a new perspective on tackling text classification problems in semi-supervised settings. Thus, in contrast to these tasks, we are not concerned about entity extraction or relation classification, but our focus is on how to decompose the classification problem such that the resulting pieces are good representations.

Another related topic, which has inspired our work, is Annotator Rationale technique introduced in \cite{Rationales}. The authors use manual annotations within documents to derive new training examples. To take into account the possible biases in the synthesized examples, they also adjust the classification model accordingly. Similar to their approach, our model also relies on the annotations within each document. The manual annotation of the sentences within each document raises efficiency concerns about the cost of preparing the training data. However, they carry out a set of extensive experiments and show that the effort of labeling the sentences within each document is not significant. Specifically, they show that when the classification task is predetermined but the set of candidate sentences and words is open and unknown, human annotators can rapidly scan the text and highlight the important sections. In our model, this issue is even less concerning, because once the set of Key Concept Sets is defined, they can be automatically \textit{discovered and highlighted}; and ready to \textit{annotate}. The main difference between \METHOD and Annotator Rationale is that our model relies on domain-guided problem decompositions to derive new training examples. Consequently, \METHOD is able to divide the initial problem into potentially smaller tasks, and tackle each one individually.

In the context of the personal health event detection, the closest work to ours is the WESPAD model introduced in \cite{PHM}--We have included the model as a baseline. The underlying assumption of WESPAD is that there is enough data to extract good lexical features. Even though this model works well in supervised settings, in Section \ref{sec:result} we will show that it performs poorly in semi-supervised settings. Finally, in contrast to general semi-supervised learning models such as transductive \cite{transductive}, graph-based \cite{graphbased}, generative \cite{EM}, or hybrid models \cite{mixmatch}, our model is a novel method to incorporate domain knowledge into the learning process. Therefore, our solution can be still implemented in any of the machine learning frameworks which can regulate the interaction between multiple learners, e.g., \cite{cotrain, DataProg, WDN}. In summary, our work advances the state of the art by identifying the problem decomposition in text classification tasks, proposing an effective co-training model to utilize the technique, and showing the superiority of the model in semi-supervised settings across multiple tasks.

\section{\METHOD\!: Method Description} \label{sec:method}

\begin{figure}
\includegraphics[scale=0.37]{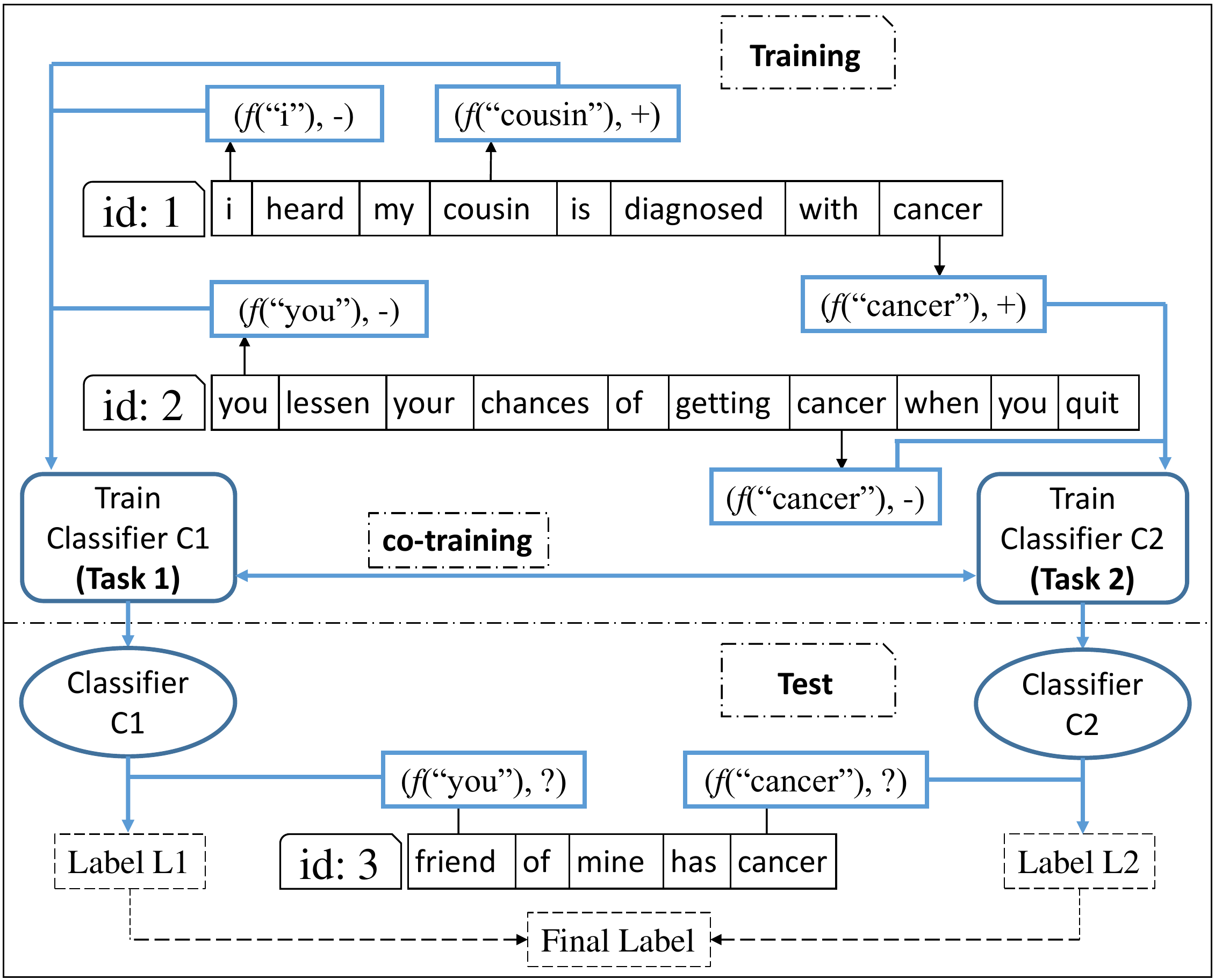}
\caption{Illustration of \METHOD method for detecting personal health mentions (cancer), where the task is decomposed into detecting positive human mentions (Class C1) and actual health event (cancer) mentions (Class C2). In the training phase, classifiers for C1 and C2 are trained over the labeled instances of C1 and C2. To label the unseen examples in the test phase, the predictions of classifiers for C1 and C2 are aggregated.} \label{fig:example}
\end{figure}

We begin this section by presenting an example, and explaining the intuition behind \METHOD\!. Consider the task of cancer surveillance in Twitter. The common practice is to extract a set of feature vectors from user postings--manually or automatically--and train a classifier over the extracted vectors. However, this approach has some drawbacks. First, the classifier needs to learn a mapping function from the linguistic patterns that appear in tweets to the class labels. Even if the patterns are not semantically and directly related to the task, the classifier still needs to learn to discard them. Second, no domain understanding is used to tackle the problem. With sufficient training data, classifiers can ultimately discover the right feature set, and detect the correct mapping function. But this is not the case in semi-supervised settings with insufficient labels. To address these issues, our proposal is to decompose the task into a set of complementary sub-tasks, and tackle each one individually.

For instance, in the case of cancer surveillance, as shown in Figure \ref{fig:example}, the original task can be decomposed into (1) detecting positive mentions of humans (marked by ``Task 1'' in Figure \ref{fig:example}) and (2) detecting positive mentions of the word cancer (marked by ``Task 2'' in Figure \ref{fig:example}). A tweet may contain multiple human mentions and cancer mentions, as shown in the case of the tweet ``id: 1'' in Figure \ref{fig:example}. The mentions that refer to the human with the reported cancer are labeled positive, while the remaining mentions are labeled as negative. Two separate classifiers are trained over the mentions of humans and the mentions of cancer, respectively. The two classifiers are then aggregated in a co-training framework to result a robust model. In the following subsections, we define Key Concept Sets and decomposable problems. Then, we describe our model \METHOD\!, which utilizes the problem decomposition in a co-training framework.

\subsection{Decomposable Text Classification Tasks} \label{subsec:method-def}
In this section, we introduce Key Concept Sets, which allow us to decompose a problem into a set of sub-tasks. Let $\pi$ be the distribution over document and class pairs $\pi{:}~(d,c) \in D \times \{0,1,\cdots\}$, and $V$ be the vocabulary set. Also let $f:(w,d,i)\mapsto\mathbb{R^n}$ be a vector-valued function which captures the contextual information of the \textit{i-th} occurrence of term $w$ in document $d$, and maps it into an $n$-dimension space of real values. Given threshold $\gamma$, we define $K$ to be a Key Concept Set if: 
\textbf{1)}~$K \subseteq V$ 
\textbf{2)}~$\forall w,v \in K{:}~{\parallel} f(w,:,:) - f(v,:,:) {\parallel} \leq \gamma$
\textbf{3)}~There exists distribution $\varphi$ over the value of $f$ and class pairs $\varphi{:}~(f,c) \in f \times \{0,1,\cdots\}$ such that $\forall d \in D, \exists~w \in K, \exists~(w,d,i): (d,c_k){\sim}\pi \Leftrightarrow (f(w,d,i),c_k){\sim}\varphi $.

Thus, a Key Concept Set is a subset of the vocabulary set--attribute (1)--in which its members are contextually similar--governed by $\gamma$ in attribute (2)--and if we train a classifier on the context vectors of its members, there is at least one term in every document where its label is the same as the document label--attribute (3). We call a classification problem decomposable, if there exists at least one Key Concept Set in the vocabulary set.

Key Concept Sets simplify the classification inference, since the classification over the documents can be replaced with the classification over the key-concept-set terms in the documents. More specifically the advantages are: First, the dimension of the context function $f$ is usually much smaller than the size of the vocabulary set $V$, thus feature selection becomes easier. Second, since intuitively there are limited ways of using a word in context, there is less variance in distribution $\varphi$ in comparison to distribution $\pi$, which can virtually model the entire language. Third, as we will discuss in the next section, we can rely on our domain understanding to identify Key Concept Sets, and therefore, equip the model with a knowledge that otherwise it would need to learn through more training data. This will help the model to generalize better with smaller number of training examples.

\subsection{Domain-Guided Key Concept Set Identification} \label{subsec:method-identify}

To identify Key Concept Sets we rely on human knowledge. Our model is proposed for the tasks which are tailored for specific entities or concepts. Therefore, we assume once the problem statement is defined, the identification of the subject entities will be straightforward. To demonstrate that this assumption holds in some real-world scenarios, in Section \ref{sec:method-app} we present three tasks that follow this motif. Namely, we discuss Personal Health Mention detection \cite{PHM}, Crisis Report detection \cite{crisis-survey}, and Adverse Drug Reaction monitoring \cite{adr} tasks. We show that, even though there is a large body of work behind each one, they can be viewed as decomposable problems and addressed similarly. This is striking, since to the best of our knowledge so far no connection has been made between these three tasks. We conjecture that there may be an even larger set of tasks that have the same attributes and can be potentially decomposable--one particularly interesting case which we may explore in the future is the product review task in social media. 

\textbf{A short note on the role of human knowledge in our model.} Our model is not a human-in-the-loop algorithm. Once the training stage begins, no human supervision is required. In the regular learning, the learner mines the entire feature space to detect the conclusive subset of features. To do so, the model requires enough training data. We are in fact eliminating this step, and reducing document level classification to word level classification. In other words, we rely on human knowledge to relocate one of the data exploration steps from the learning stage to the design stage. Thus, the learning procedure still occurs, however, in a smaller feature space with less variation. The idea of reliance on human knowledge is not novel. For instance, the distant supervision model \cite{distantsup}, assumes the user has enough domain expertise to introduce a large noisy dataset. Co-training model  \cite{cotrain}, assumes the user has enough information about the task to introduce two subsets of features. And the data programming model \cite{DataProg}, assumes the user has enough knowledge to provide the learner with a set of heuristics. Interestingly, all of these models are proposed for the low data regime.

\subsection{\METHOD\!: Exploiting Task Decomposition for Semi-Supervised Learning} \label{subsec:method-impl}


The contextual similarity between the members of a Key Concept Set, that was introduced in the previous section, insures that the sets that can potentially capture different aspects of documents are not combined \footnote{The similarity condition--introduced by $\gamma$--does not by itself guarantee orthogonality of the features. However, if two subsets of vocabularies are contextually different, and their context vectors are indicators of the document class, then, we assume they can capture different aspects of the document.}. Being able to capture multiple views of the same problem--even loosely--is shown to be effective in models such as co-training \cite{cotrain, cotrain-2}. Thus, we propose to use co-training to utilize the problem decomposition\footnote{We consider the binary classification problems, however, our model can also generalize to multi-label classification problems.}. Algorithm \ref{alg:train} illustrates the training procedure of \METHOD\!. Since there could be multiple occurrences of the members of a Key Concept Set in a document, the problem is viewed as a multiple instance learning problem \cite{MIL}, where each document is called an example, and each set member occurrence in the document is called an instance. The procedure is iterative, and in every iteration the set of labeled instances of every example are used to train a classifier. Then the classifiers are used to label the instances of the unlabeled data, and according to the multiple instance learning selection metric the examples are labeled--e.g, based on their most confident positive instance. Finally, the most confident positive and negative examples of each Key Concept Set are added to the pool of the labeled training data.


\begin{algorithm}
\caption{Training Procedure of \METHOD}\label{alg:train}
\begin{algorithmic}[1]
\algrenewcommand\algorithmicindent{7pt}
\Procedure{Train}{}
    \State \textbf{Given:}
    \Indent
        \State $L: \text{Set of labeled examples}$ 
        \State $U: \text{Set of unlabeled examples}$
        \State $J: \text{Number of key concept sets}$
        \State $K: \text{Number of iterations}$
    \EndIndent
    \State \textbf{Return:}
    \Indent
        \State $C[1\dotsc J]:$ \parbox[t]{.7\linewidth}{array of classifiers trained on instances of each key concept set in $L$ and $U$}
    \EndIndent
    \State \textbf{Execute:}
    \Indent
        \For{$i \gets 1$ to $K$}
            \For{$j \gets 1$ to $J$}
                \State \text{Train $C_j$ on \textbf{instances} of key concept set $j$ in $L$}
                \State \parbox[t]{.8\linewidth}{Use $C_j$ and multiple instance learning metric to label the \textbf{examples} in $U$}
                \Indent
                    \State \parbox[t]{.8\linewidth}{Store the most confident positive and negative \textbf{examples} in $EP_j$ and $EN_j$}
                \EndIndent
            \EndFor
            \For{$j \gets 1$ to $J$}
                \State \text{Delete $EP_j$ and $EN_j$ in $U$ and add them to $L$}
            \EndFor
        \EndFor
        \State \text{Return $C[1\dotsc J]$}
    \EndIndent
\EndProcedure
\end{algorithmic}
\end{algorithm}

Algorithm \ref{alg:test} illustrates the test procedure. The array of classifiers trained in Algorithm \ref{alg:train} are used to label the unseen examples. To label every example, each classifier is used to calculate the probability of the example being positive, and then a simple criterion similar to the one proposed in \cite{cotrain} is used to label the example. In a more complicated scenario, each classifier could have a prior reliability score, however, for simplicity we opted for the model proposed in  \cite{cotrain}.

\begin{algorithm}
\caption{Test Procedure of \METHOD}\label{alg:test}
\begin{algorithmic}[1]
\algrenewcommand\algorithmicindent{5pt}
\Procedure{Test}{}
    \State \textbf{Given:}
    \Indent
        \State $J: \text{Number of key concept sets}$
        \State $C[1\dotsc J]: \text{array of classifiers}$
        \State $Test: \text{Test set}$
    \EndIndent
    \State \textbf{Return:}
    \Indent
        \State \text{Labeled test set}
    \EndIndent
    \State \textbf{Execute:}
    \Indent
        \For{$exmpl~\textbf{in}~Test$}
            \For{$j \gets 1$ to $J$}
                \State \parbox[t]{.9\linewidth}{Use $C_j$ and multiple instance learning metric to find the probability of $exmpl$ being positive}
                \Indent
                    \State \text{Store the corresponding probability in $P_j$}
                \EndIndent
            \EndFor
            \If{$\prod_{i=1}^{J} P_i \geq \prod_{i=1}^{J} (1 - P_i)$}
                \State $exmpl$ is positive
            \Else
                \State $exmpl$ is negative
            \EndIf
        \EndFor
        \State \text{Return $Test$}
    \EndIndent
\EndProcedure
\end{algorithmic}
\end{algorithm}

\textbf{A short note on the orthogonality of Key Concept Sets.} Multi-view learning techniques \cite{multi-view} are effective even in the presence of correlated views. Particularly in the case of co-training algorithm, numerous studies have shown that the initial assumption of orthogonality between the views was over-strong. For instance, Balcan, Blum, and Yang \cite{co-traing-analysis} propose a theoretical framework and argue that if the classifiers in each view are sufficiently strong PAC-learners, then the initial constraint on the views can be substantially relaxed. In the application domain, Nigam and Ghani \cite{cotrain-2} show that by randomly splitting lexical features, one can construct two separate views for co-training algorithm. Jones et al., \cite{co-em}, propose Co-EM algorithm for information extraction. Their two feature sets are noun phrases and their surrounding contexts. They show that even though these two feature sets are highly correlated, they can be still effective in a co-training model.

In the next section, we use \METHOD to propose a solution to a set of personal event detection tasks in social media.

\section{Applications: Personal Event Detection} \label{sec:method-app}

In this section, we show that \METHOD is applicable to three important real-world scenarios: Personal Health Mention detection (PHM), Crisis Report detection (CR), and Adverse Drug Reaction monitoring (ADR). We show that these three tasks are decomposable problems and have a unified solution.

\subsection{Personal Health Mention Detection} \label{subsec:method-phm}

Personal Health Mention detection (PHM) is described in \cite{PHM}, and concerns \textit{``identifying postings in social data, which not only contain a specific \textbf{disease}, but also mention a \textbf{person} who is affected''}. To employ \METHOD\!, we regard the two entities that are present in the problem statement as the Key Concept Sets: 1) The set of all human mentions. 2) The disease keyword mentioned in the task. We argue that both of the sets loosely follow the conditions which are described in Section \ref{subsec:method-def}. Intuitively, all the human mentions have similar contextual vectors (condition (2)); and by construction, there is at least one human mention that determines the label of the user posting (condition (3)). The same reasoning applies to the second Key Concept Set; there must be at least one occurrence of the disease keyword which determines the label of the user posting (condition~(3)). 



After identifying the Key Concept Sets, the next step is to prepare the training set. We implemented a tool to automatically extract the human mentions and highlight the mentions for manual annotation--similar to Annotators Rationale method \cite{Rationales}. Since user postings are short, we assumed all the disease mentions in the positive user postings were positive instances of the second Key Concept Set. All the human mentions and disease mentions of the negative user postings were assumed to be negative instances. Thus, the extraction and annotation of the disease mentions, the extraction of the human mentions, and also the annotation of the negative human mentions are all fully automatic. Only the annotation of the positive human mentions is manual--after a tweet is labeled positive, the user is asked to highlight the affected human mention.

We followed Algorithm~\ref{alg:train} for training the classifiers, and augmented the labeled data with unlabeled data. To add positive instances of Key Concept Sets to the labeled data, we selected the most confidently labeled instance and its most probable counterpart in the other Key Concept Set--we effectively stored the set of instances as labeled data. For example, assume the classifier trained over disease mentions confidently labeled the word ``cancer'' positive in the tweet \textit{``a friend of me is diagnosed with cancer''}. Then, we added this instance to the set of labeled data, and also used the classifier trained over the human mentions to label the mentions of human in the tweet, i.e., ``friend'' and ``me'', and selected the most confident one and added to the labeled data. To add negative instances of Key Concept Sets to the labeled data, we selected the example which all of its instances were confidently labeled negative, and added to the labeled data. To test our model, we followed Algorithm~\ref{alg:test}.

\subsection{Crisis Report Detection} \label{subsec:method-cr}

Crisis Report detection (CR) as defined in \cite{crisis-dataset} concerns\footnote{There are also other variations of this task, e.g., displacing or evacuating people, during different incidents  \cite{crisis-semi}.} \textit{``detecting reports of casualties and/or injured \textbf{people} due to the \textbf{crisis}. Or reports and/or questions about missing or found \textbf{people}''}. We regard the two entities mentioned in the problem statement as the Key Concept Sets: 1) The set of all human mentions. 2) The crisis keyword mentioned in the task. In this study, we focus on the reports which were posted during an earthquake. To prepare the training set and evaluate our model, we followed the same procedure that we used for the PHM problem.

\subsection{Adverse Drug Reaction Monitoring} \label{subsec:method-adr}
Adverse Drug Reaction monitoring (ADR) is defined in \cite{adr2}, and is meant for \textit{``detecting \textbf{personal} injuries resulting from medical \textbf{drug} use''}. We regard the two entities mentioned in the problem statement as the Key Concept Sets: 1) The set of all human mentions. 2) The set of all drug mentions. To prepare the training set and evaluate our model, we re-implemented all the decisions that we made for the PHM problem.

\subsection{Implementation Details} \label{subsec:method-case-impl}
In this section we provide a detailed explanation of the modules and components used in \METHOD to address the tasks mentioned earlier. Specifically, we discuss the context function described in Section \ref{subsec:method-def}, the classifiers described in Section \ref{subsec:method-impl}, the extraction of the Key Concept Sets mentioned in Sections \ref{subsec:method-phm}, \ref{subsec:method-cr}, and \ref{subsec:method-adr}; and finally the learning representation of the Key Concept Sets.

\noindent\textbf{Context Function}. We used contextual embeddings as the context function described in Section \ref{subsec:method-def}. We used the BERT model \cite{bert}, even though other models such as ELMO could be also used \cite{elmo}. We used the base variant, and pre-trained it on Twitter data--see below for the details about pre-training.

\noindent\textbf{Used Classifiers}. We used logistic regression classifier as the learners mentioned in Section \ref{subsec:method-impl}. Thus, after fine-tuning the embeddings using the training data, we used the contextual features to train the logistic regression classifiers\footnote{We made this decision based on implementation considerations.}. The Mallet implementation of logistic regression \cite{mallet} was used in this step.

\noindent\textbf{Key Concept Set Extraction}. To detect human mentions we used a weak rule-based classifier. The accurate detection of human mentions is out of our research scope; here, we aim to show that even a weak human mention detector can contribute to the performance. The rules for human mention detection were as follows: Using the Stanford Named Entity Recognition (NER) tagger \cite{stanfordner} we labeled all of the ``PERSON'' tags. Using the Stanford Parts of Speech (POS) tagger \cite{stanfordpos} we labeled all of the personal pronoun tags except for the word ``it''. We also labeled all of the Twitter mentions--indicated by the sign ``@''. Finally, we used a dictionary of 240 words manually collected from the Web to cover the remaining cases. Since not all of the human mentions are explicitly referred in user postings, we also used a simple noisy rule based human mention synthesizer: If a sentence started with a past tense verb we inserted the word ``i'' at the beginning. If a sentence started with an adjective we inserted ``i am'' at the beginning. If a sentence started with a past participle verb we inserted ``i have'' at the beginning. If a sentence started with a present continuous verb we inserted ``i am'' at the beginning. And finally, if a sentence started with ``is'', we replaced it with ``i am''. We empirically developed these rules, and as mentioned earlier, to achieve a better performance they can be replaced with more sophisticated models.

The model relies on the positive mentions of the humans in the positive tweets--described in Section \ref{subsec:method-phm}. One of the authors of the article supplied the annotations. The rules for the annotation were as follows: The explicit mentions of the humans which are associated with the event (either disease, or disaster, or drug injury) should be annotated. If the explicit mention does not exist, the implicit mentions which are associated with the event should be annotated.

To extract the disease Key Concept Set mentioned in Section \ref{subsec:method-phm}, we conducted a keyword search for the disease name in the task description. For instance if the task is about Parkinson's disease surveillance, the disease Key Concept Set contains the word \{Parkinson's\}. To extract the crisis Key Concept Set mentioned in Section \ref{subsec:method-cr}, we also performed a keyword search for the incident in the task description. As mentioned earlier, in this study we focused on an earthquake incident. Thus, the crisis Key Concept Set contains the keywords \{earthquake, quake\}. To extract the drug Key Concept Set described in Section \ref{subsec:method-adr}, we used the list of drug names published in \cite{adr}, and conducted a keyword search for the drug names in the list.

\noindent\textbf{Learning Key Concept Set Representations}. Since the human mentions are lexically different--although we expect them to be contextually similar--we replaced all of them with a mask token HUM\_TOK and learned the representation. To do so, we collected a set of 7,598,545 random tweets by Twitter API in October 2018, replaced all the human mentions with this token, and pre-trained the base variant of the BERT model for 10 epochs--with default hyperparameters as mentioned in \cite{bert}. The word vectors used in the personal health mention detection and crisis report detection tasks are the output of this model. To unify the representations of the drug mentions, we used the list of drug names published in \cite{adr} to collect a set of 28,710 tweets containing the drug names~\footnote{We used the Twitter streaming API for four weeks, and collected about 300K tweets, however, found that the majority of them were duplicates.}, replaced the names with DRUG\_TOK and further pre-trained the above mentioned model for 10 epochs. The word vectors used in the adverse drug reaction monitoring task are the output of this model.



\section{Experimental Setup} \label{sec:setup}

In this section we first describe the datasets that we used in the experiments, and then, we review the baselines that we implemented, and finally discuss the training procedure. 

\subsection{Datasets} \label{subsec:dataset}

For personal health mention detection task we used two datasets. First, the dataset introduced in \cite{flu-data}, which we call \FLUDT dataset~\footnote{We used the infection vs awareness version of \FLUDT dataset, for detailed information about the datasets please refer to the cited articles.}. At the time of downloading this dataset, there were still 2,837 tweets available to crawl, in which 49\% of them are negative--awareness tweets--and 51\% of them are positive--report actual cases of flu. Second, the dataset introduced in \cite{PHM}, which we call \PHMDT dataset. At the time of downloading this dataset, there were 7,192 tweets available to crawl. This dataset consists of 6 diseases: Alzheimer's, heart attack, Parkinson's, cancer, depression, and stroke. All of these sub-datasets are highly imbalanced, positive examples span between 11\% to 40\% of the cases. For crisis report detection task, we used the earthquake related dataset introduced in \cite{crisis-dataset}, which we call \CRISISDT dataset. This dataset contains a set of 2,013 tweets which were posted during the California earthquake in 2014~\footnote{Reference \cite{crisis-dataset} also introduces a few more datasets. We used the California earthquake version, and split by the injured and missing vs other categories.}. Only 11\% of the tweets in this dataset are positive cases of injured or missing people. For adverse drug reaction monitoring task, we used the dataset introduced in \cite{adr}, which we call \ADRDT dataset. At the time of crawling the dataset, there were 4,355 tweets available. This dataset is also highly imbalanced, only 10\% of the tweets are positive cases of drug injures. Table \ref{tab:dataset} summarizes the 4 datasets and their target prediction tasks.

\begin{table}
\begin{tabular}{|p{0.6in} |p{1in} |p{0.5in} |p{0.6in} |}\hline
\textbf{Name} & 
\textbf{Target} & 
\textbf{\# Tweets} & 
\textbf{\% Positive} \\ \hline
\FLUDT \cite{flu-data} & Positive flu cases & 2837 & 51 \\ \hline 
\PHMDT \cite{PHM} & Alzheimer & 1256 & 18 \\ \hline 
\PHMDT \cite{PHM} & Heart attack & 1219 & 13 \\ \hline 
\PHMDT \cite{PHM} & Parkinson's & 1040 & 11 \\ \hline 
\PHMDT \cite{PHM} & Cancer & 1242 & 21 \\ \hline 
\PHMDT \cite{PHM} & Depression & 1213 & 40 \\ \hline 
\PHMDT \cite{PHM} & Stroke & 1222 & 14 \\ \hline 
\CRISISDT \cite{crisis-dataset} & Injured or missing & 2013 & 11 \\ \hline 
\ADRDT \cite{adr}& Drug injuries & 4355 & 10 \\ \hline 
\end{tabular}
\caption{Summary of \FLUDT \cite{flu-data}, \PHMDT \cite{PHM}, \CRISISDT \cite{crisis-dataset}, and \ADRDT \cite{adr} datasets and their associated prediction tasks. The third and fourth columns report the size of the dataset and percentage of the positive tweets respectively.} \label{tab:dataset}
\vspace{-0.5cm}
\end{table}

\subsection{Baselines} \label{subsec:baselines}

To compare the performance of our method, we implemented the following methods and classifiers. Model hyperparameters were tuned based on the training folds and datasets, and in most cases their optimal values were dependent on the training data.

\noindent\textbf{\textit{NB}}. A Naive Bayes classifier is trained over unigrams and bigrams, as it has been shown to perform well with small training sets \cite{NB}.

\noindent\textbf{\textit{EM}}. We implemented the Expectation Maximization algorithm proposed by \cite{EM}, which is known to work well in semi-supervised settings. We experimented with the set of \{10,20,50,100\} for the number of unlabeled documents.

\noindent\textbf{\textit{FastText}}. We trained the shallow neural network classifier introduced in \cite{fasttext}, which can update word embeddings during the training. We experimented with \{0.05,0.1,0.25,0.5\} for the learning rate, and \{2,4\} for the window size.

\noindent\textbf{\textit{WESPAD}}. We trained the PHM model introduced in \cite{PHM}, which is specifically designed for Personal Health Mention detection. We experimented with \{3,4,5\} for the number of clusters, and \{0.05,0.15,0.3\} for threshold values.

\noindent\textbf{\textit{BERT-BASE}}. We included the model introduced in \cite{bert}, which is named BERT and uses a multi-layer transformer encoder followed by one layer of a fully connected neural network for binary classification problems. In the experiments we observed that the large variant shows poor performance when the training data is small, thus we report the results of the base variant \textit{BERT-BASE}--which has fewer layers. We followed the parameter settings suggested in \cite{bert}; but empirically observed that if we set the number of epochs for fine-tuning to 15, the model is more stable and performs better.

\noindent\textbf{\textit{BERT-TW}}. Since we experimented with Twitter data, we also pre-trained BERT in order to adjust the language model. Thus, we used the set of 7 million tweets described in Section \ref{subsec:method-case-impl} to further pre-train \textit{BERT-BASE} for 10 epochs--without replacing human mentions. The hyperparameters were set to what is suggested in \cite{bert}, and by the time the pre-training was done, the performance of the internal language modelling tasks for sample tweets was similar to the performance of \textit{BERT-BASE} for sample Wikipedia pages.

\noindent\textbf{\textit{BERT-DR}}. We also used the set of drug related tweets mentioned in Section \ref{subsec:method-case-impl}--without replacing the drug mentions--to further pre-train \textit{BERT-TW} to be used in ADR task. We used the same parameter setting as \textit{BERT-TW}.

\noindent\textbf{\textit{Co-BE-LE}}. In order to boost the BERT model with Bootstrapping, we also included a \emph{co-training model} with two learners: One Naive Bayes classifier trained over unigrams and bigrams, and one logistic regression classifier trained over the \textit{BERT-TW} or \textit{BERT-DR} representation of the tweets--depending on the task. We experimented with \{13,25,50\} as the number of iterations in co-training model.

\noindent\textbf{\textit{\METHOD}\!}. Our method described in Section \ref{sec:method-app}. We empirically set the number of iterations in the co-training model to 25--based on the training and development folds in the \FLUDT dataset--and did not do any further tuning beyond what we did for \textit{BERT-TW}. We report all the results with this setting unless stated otherwise.

\subsection{Training Details} \label{subsec:train-detail}

We used standard 10-fold cross validation to train, validate, and test all of the models. To evaluate the models in semi-supervised settings, we did not use the entire training and validation data, but randomly sampled a few examples and used the rest of the examples as unlabeled data. In the next section, we report the results when we have 100 training examples, however, we also show that our model still performs well when the number of available training examples increases. To split the datasets into the folds, we used stratified sampling to preserve the original class distribution in the datasets. We also preserved the folds and samples identical across the experiments to ensure that all of the models use exactly the same training and test data. Since there is a natural randomness in neural network initialization and regularization techniques, we carried out all of the experiments 5 times, and averaged the performance results.

Because the datasets are highly imbalanced, following the argument in \cite{perf-metric}, we used the F1 measure in the positive class to tune the models. In the next section we report F1, Precision, and Recall in the positive class--averaged over the test folds.

\section{Results and Discussion} \label{sec:result}

In this section, we first report the performance results in \FLUDT\!, \PHMDT\!, \CRISISDT\!, and \ADRDT datasets, and then  analyze our model through a series of experiments.

\subsection{Performance Results} \label{subsec:result-short}

Table \ref{tbl-result-phm} summarizes the F1, precision, and recall of the models in \FLUDT and \PHMDT datasets--the results in \PHMDT dataset are averaged over the topics. Table \ref{tbl-result-crisis} summarizes the results in \CRISISDT dataset, and Table \ref{tbl-result-adr} reports the results in \ADRDT dataset. We also report the performance of the models in \PHMDT dataset across all the topics in Table \ref{tbl-result-phm-details}. The experiments show that \METHOD outperforms state-of-the-art classifiers across the majority of the tasks. We can see that the improvements in the imbalanced datasets (\PHMDT and \ADRDT\!) are more noticeable than the improvements in the balanced dataset (\FLUDT\!). We can also see that the semi-supervised learning model \textit{Co-BE-LE} performs relatively well, although it has a low precision. In contrast, our model maintains a high precision. We attribute this advantage to the easier tasks that \METHOD is tackling--i.e., selecting the most confident unlabeled instances via the context representations versus via the document representations. Finally, the results suggest that crisis report detection is an easier problem than adverse drug reaction monitoring, because even though both \CRISISDT and \ADRDT have about 10\% positive examples, the performance of the models in the \ADRDT dataset is much lower. We will discuss this dataset in more detail in the next section.


\begin{table*}
\begin{tabu}{p{0.8in} |p{0.5in} |p{0.5in} |p{0.5in} |[1pt] p{0.5in} |p{0.5in} |p{0.5in} |}
\cline{2-7} & \multicolumn{3}{|c|[1pt]}{{\footnotesize \textbf{\FLUDT dataset}}}  &
 \multicolumn{3}{|c|}{{\footnotesize \textbf{\PHMDT dataset}}}  \\ \hline 
\multicolumn{1}{|c|}{\textbf{Model}} & 
{\footnotesize \textbf{F1}} & 
{\footnotesize \textbf{Precision}} & 
{\footnotesize \textbf{Recall}} &
{\footnotesize \textbf{F1}} & 
{\footnotesize \textbf{Precision}} & 
{\footnotesize \textbf{Recall}} \\ \hline
\multicolumn{1}{|c|}{\textit{NB}} & 0.752 & 0.712 & 0.800 & 0.304 & 0.616 & 0.255 \\ \hline 
\multicolumn{1}{|c|}{\textit{EM}} & 0.766 & 0.708 & \textbf{0.843} & 0.407 & 0.528 & 0.414 \\ \hline 
\multicolumn{1}{|c|}{\textit{FastText}} & 0.747 & 0.728 & 0.772 & 0.278 & 0.626 & 0.215 \\ \hline 
\multicolumn{1}{|c|}{\textit{WESPAD}} & 0.763 & 0.728 & 0.805 & 0.336 & 0.668 & 0.272 \\ \hline 
\multicolumn{1}{|c|}{\textit{BERT-BASE}} & 0.757 & 0.739 & 0.790 & 0.572 & 0.682 & 0.537 \\ \hline 
\multicolumn{1}{|c|}{\textit{BERT-TW}} & 0.786 & 0.782 & 0.800 & 0.563 & \textbf{0.698} & 0.512 \\ \hline 
\multicolumn{1}{|c|}{\textit{Co-BE-LE}} & 0.771 & 0.715 & 0.838 & 0.577 & 0.627 & 0.593 \\ \hline 
\multicolumn{1}{|c|}{\textit{\METHOD}} & \textbf{0.809} & \textbf{0.800} & 0.822 & \textbf{0.630} & 0.674 & \textbf{0.617} \\ \hline 
\end{tabu}
\caption{F1, precision, and recall in \FLUDT\! and \PHMDT\! datasets for all the models.
} \label{tbl-result-phm}
\vspace{-0.5cm}
\end{table*}

\begin{table}
\begin{tabu}{p{0.75in} | p{0.5in} |p{0.5in} |p{0.5in} |}
\cline{2-4} &  
 \multicolumn{3}{|c|}{{\footnotesize \textbf{\CRISISDT dataset}}} \\ \hline 
\multicolumn{1}{|c|}{\textbf{Model}} & 
{\footnotesize \textbf{F1}} & 
{\footnotesize \textbf{Precision}} & 
{\footnotesize \textbf{Recall}} \\ \hline
\multicolumn{1}{|c|}{\textit{NB}} & 0.545 & 0.865 & 0.400 \\ \hline 
\multicolumn{1}{|c|}{\textit{EM}} & 0.568 & 0.625 & 0.535 \\ \hline 
\multicolumn{1}{|c|}{\textit{FastText}} & 0.382 & 0.815 & 0.258 \\ \hline 
\multicolumn{1}{|c|}{\textit{WESPAD}} & 0.607 & \textbf{0.932} & 0.458 \\ \hline 
\multicolumn{1}{|c|}{\textit{BERT-BASE}} & 0.710 & 0.818 & 0.676 \\ \hline 
\multicolumn{1}{|c|}{\textit{BERT-TW}} & 0.732 & 0.859 & 0.678 \\ \hline 
\multicolumn{1}{|c|}{\textit{Co-BE-LE}} & 0.609 & 0.615 & 0.614 \\ \hline 
\multicolumn{1}{|c|}{\textit{\METHOD}} & \textbf{0.765} & 0.880 & \textbf{0.694} \\ \hline 
\end{tabu}
\caption{F1, precision, and recall in \CRISISDT dataset for all the models.
} \label{tbl-result-crisis}
\vspace{-0.5cm}
\end{table}

\begin{table}
\begin{tabu}{p{0.75in} | p{0.5in} |p{0.5in} |p{0.5in} |}
\cline{2-4} &  
 \multicolumn{3}{|c|}{{\footnotesize \textbf{\ADRDT dataset}}} \\ \hline 
\multicolumn{1}{|c|}{\textbf{Model}} & 
{\footnotesize \textbf{F1}} & 
{\footnotesize \textbf{Precision}} & 
{\footnotesize \textbf{Recall}} \\ \hline
\multicolumn{1}{|c|}{\textit{NB}} & 0.020 & 0.267 & 0.011 \\ \hline 
\multicolumn{1}{|c|}{\textit{EM}} & 0.072 & 0.168 & 0.052 \\ \hline 
\multicolumn{1}{|c|}{\textit{FastText}} & 0.004 & 0.100 & 0.002 \\ \hline 
\multicolumn{1}{|c|}{\textit{WESPAD}} & 0.016 & 0.300 & 0.008 \\ \hline 
\multicolumn{1}{|c|}{\textit{BERT-BASE}} & 0.082 & 0.274 & 0.054 \\ \hline 
\multicolumn{1}{|c|}{\textit{BERT-DR}} & 0.098 & 0.290 & 0.066 \\ \hline 
\multicolumn{1}{|c|}{\textit{Co-BE-LE}} & 0.184 & 0.183 & 0.202 \\ \hline 
\multicolumn{1}{|c|}{\textit{\METHOD}} & \textbf{0.259} & \textbf{0.302} & \textbf{0.236} \\ \hline 
\end{tabu}
\caption{F1, precision, and recall in \ADRDT dataset for all the models.
} \label{tbl-result-adr}
\vspace{-0.5cm}
\end{table}

\begin{table*}
\begin{tabu}{p{0.8in} |p{0.5in} |p{0.6in} |p{0.6in} |[1pt] p{0.5in} |p{0.6in} |p{0.6in} |}

\cline{2-7} & \multicolumn{3}{c|[1pt]}{{\footnotesize \textbf{Alzheimer's}}}  &
 \multicolumn{3}{c|}{{\footnotesize \textbf{Heart attack}}}  \\ \hline 
\multicolumn{1}{|c|}{\textbf{Model}} & 
{\footnotesize \textbf{F1}} & 
{\footnotesize \textbf{Precision}} & 
{\footnotesize \textbf{Recall}} &
{\footnotesize \textbf{F1}} & 
{\footnotesize \textbf{Precision}} & 
{\footnotesize \textbf{Recall}} \\ \hline
\multicolumn{1}{|c|}{\textit{NB}} & 0.534 & 0.859 & 0.403 & 0.058 & 0.400 & 0.032 \\ \hline 
\multicolumn{1}{|c|}{\textit{EM}} & 0.617 & 0.663 & 0.618 & 0.072 & 0.500 & 0.039 \\ \hline 
\multicolumn{1}{|c|}{\textit{FastText}} & 0.418 & \textbf{0.890} & 0.284 & 0.048 & 0.400 & 0.025 \\ \hline 
\multicolumn{1}{|c|}{\textit{WESPAD}} & 0.535 & 0.837 & 0.421 & 0.058 & 0.400 & 0.032 \\ \hline 
\multicolumn{1}{|c|}{\textit{BERT-BASE}} & \textbf{0.698} & 0.723 & 0.701 & 0.366 & 0.586 & 0.309 \\ \hline 
\multicolumn{1}{|c|}{\textit{BERT-TW}} & 0.660 & 0.728 & 0.634 & 0.425 & 0.675 & 0.332 \\ \hline 
\multicolumn{1}{|c|}{\textit{Co-BE-LE}} & 0.682 & 0.674 & \textbf{0.721} & 0.378 & 0.647 & 0.298 \\ \hline 
\multicolumn{1}{|c|}{\textit{\METHOD}} & 0.676 & 0.694 & 0.682 & \textbf{0.534} & \textbf{0.684} & \textbf{0.451} \\ \hline 

\cline{1-7} & \multicolumn{3}{c|[1pt]}{{\footnotesize \textbf{Parkinson's}}}  &
 \multicolumn{3}{c|}{{\footnotesize \textbf{Cancer}}}  \\ \hline 
\multicolumn{1}{|c|}{\textbf{Model}} & 
{\footnotesize \textbf{F1}} & 
{\footnotesize \textbf{Precision}} & 
{\footnotesize \textbf{Recall}} &
{\footnotesize \textbf{F1}} & 
{\footnotesize \textbf{Precision}} & 
{\footnotesize \textbf{Recall}} \\ \hline
\multicolumn{1}{|c|}{\textit{NB}} & 0.155 & 0.563 & 0.096 & 0.278 & 0.661 & 0.181 \\ \hline 
\multicolumn{1}{|c|}{\textit{EM}} & 0.356 & 0.521 & 0.301 & 0.429 & 0.492 & 0.424 \\ \hline 
\multicolumn{1}{|c|}{\textit{FastText}} & 0.076 & 0.350 & 0.043 & 0.219 & \textbf{0.706} & 0.134 \\ \hline 
\multicolumn{1}{|c|}{\textit{WESPAD}} & 0.188 & \textbf{0.683} & 0.113 & 0.335 & 0.682 & 0.227 \\ \hline 
\multicolumn{1}{|c|}{\textit{BERT-BASE}} & 0.451 & 0.631 & 0.387 & 0.570 & 0.679 & 0.515 \\ \hline 
\multicolumn{1}{|c|}{\textit{BERT-TW}} & 0.452 & 0.597 & 0.405 & 0.534 & 0.700 & 0.466 \\ \hline 
\multicolumn{1}{|c|}{\textit{Co-BE-LE}} & 0.518 & 0.551 & 0.546 & 0.569 & 0.632 & 0.573 \\ \hline 
\multicolumn{1}{|c|}{\textit{\METHOD}} & \textbf{0.560} & 0.520 & \textbf{0.630} &\textbf{ 0.627} & 0.704 & \textbf{0.581} \\ \hline 

\cline{1-7} & \multicolumn{3}{c|[1pt]}{{\footnotesize \textbf{Depression}}}  &
 \multicolumn{3}{c|}{{\footnotesize \textbf{Stroke}}}  \\ \hline 
\multicolumn{1}{|c|}{\textbf{Model}} & 
{\footnotesize \textbf{F1}} & 
{\footnotesize \textbf{Precision}} & 
{\footnotesize \textbf{Recall}} &
{\footnotesize \textbf{F1}} & 
{\footnotesize \textbf{Precision}} & 
{\footnotesize \textbf{Recall}} \\ \hline
\multicolumn{1}{|c|}{\textit{NB}} & 0.670 & 0.617 & 0.742 & 0.130 & 0.597 & 0.076 \\ \hline 
\multicolumn{1}{|c|}{\textit{EM}} & 0.671 & 0.563 & \textbf{0.841} & 0.298 & 0.431 & 0.259 \\ \hline 
\multicolumn{1}{|c|}{\textit{FastText}} & 0.702 & \textbf{0.744} & 0.671 & 0.208 & 0.663 & 0.129 \\ \hline 
\multicolumn{1}{|c|}{\textit{WESPAD}} & 0.713 & 0.715 & 0.722 & 0.187 & 0.688 & 0.117 \\ \hline 
\multicolumn{1}{|c|}{\textit{BERT-BASE}} & 0.729 & 0.727 & 0.745 & 0.617 & 0.746 & 0.564 \\ \hline 
\multicolumn{1}{|c|}{\textit{BERT-TW}} & \textbf{0.737} & 0.740 & 0.747 & 0.569 & \textbf{0.752} & 0.490 \\ \hline 
\multicolumn{1}{|c|}{\textit{Co-BE-LE}} & 0.718 & 0.662 & 0.791 & 0.596 & 0.595 & 0.630 \\ \hline 
\multicolumn{1}{|c|}{\textit{\METHOD}} & 0.711 & 0.715 & 0.717 & \textbf{0.673} & 0.732 & \textbf{0.643} \\ \hline 

\end{tabu}
\caption{F1, precision, and recall of the models across the topics in \PHMDT\! dataset.} \label{tbl-result-phm-details}
\vspace{-0.5cm}
\end{table*}

\subsection{Discussion} \label{subsec:result-discuss}

To better understand the impact of each component in our model, we report the results of the ablation study in Table \ref{tbl-result-imp-detail}. Since \PHMDT dataset was the most diverse dataset (it constitutes 6 sub-topics), we carried out the experiment in this dataset. The results show that the weak human mention classifier is clearly contributing to the performance when it is combined with the disease mention classifier. Then a further improvement is achieved when co-training iterations are performed. However, the improvement after 50 iterations comes at the cost of dramatic deterioration in precision, which might not be desirable.

\begin{table}
\begin{tabular}{|p{1in} |p{0.35in} |p{0.5in} |p{0.5in} |}\hline
\textbf{Model} & 
{\footnotesize \textbf{F1}} & 
{\footnotesize \textbf{Precision}} & 
{\footnotesize \textbf{Recall}} \\ \hline
\textit{Human-cl} & 0.390 & 0.326 & 0.521 \\ \hline 
\textit{Disease-cl} & 0.541 & 0.707 & 0.469 \\ \hline 
\textit{Combined} & 0.557 & 0.733 & 0.489 \\ \hline 
\textit{+13-itr co-train} & 0.608 & 0.705 & 0.565 \\ \hline 
\textit{+25-itr co-train} & 0.630 & 0.674 & 0.617 \\ \hline 
\textit{+50-itr co-train} & 0.637 & 0.587 & 0.715 \\ \hline 
\textit{+75-itr co-train} & 0.627 & 0.545 & 0.768 \\ \hline 
\end{tabular}
\caption{Improvement analysis in \PHMDT dataset. The performance of human mention classifier (\textit{Human-cl}), disease mention classifier (\textit{Disease-cl}), their combination in co-training framework without adding unlabeled data (\textit{Combined}), and when unlabeled data is added per co-training iteration (4 unlabeled documents are added in every iteration).} \label{tbl-result-imp-detail}
\end{table}

\begin{figure}
\includegraphics[width=3.3in]{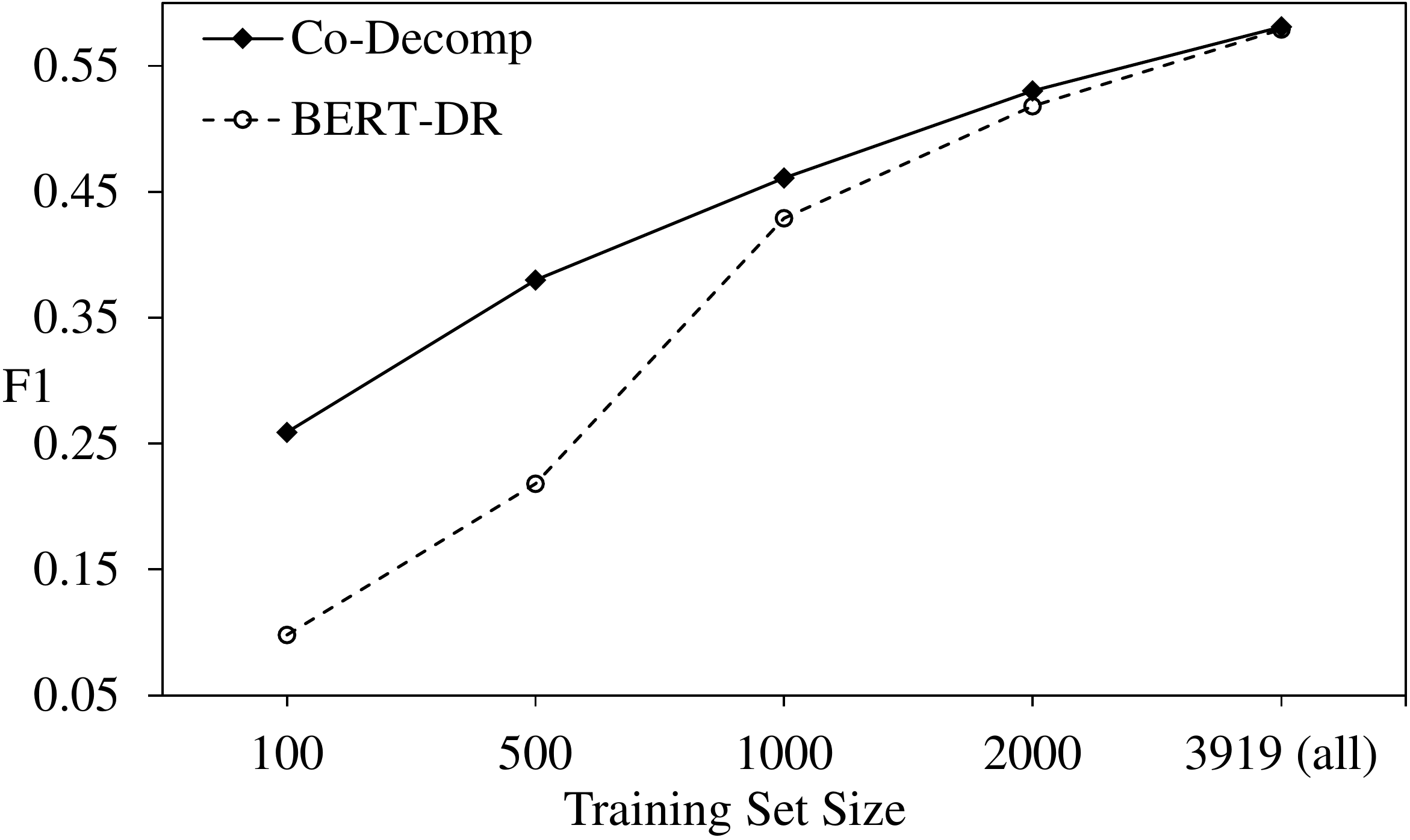}
\caption{F1 at different training set size cut-offs for \textit{BERT-DR} and \textit{\METHOD} models in \ADRDT dataset. There are 3,919 examples in the training folds of \ADRDT dataset--excluding the test folds in 10 fold cross validation.} \label{fig:result-train-size}
\end{figure}

In Section \ref{subsec:result-short}, we observed that the performance of the models in \ADRDT dataset was very low. To investigate the performance of the models as the function of the training set size, in Figure \ref{fig:result-train-size} we report the performance of \METHOD in comparison to the state-of-the-art \textit{BERT-DR} classifier at different training set size cut-offs in this dataset. The results show that even in supervised settings our model is on par with strong classifiers--for this dataset and with manual feature engineering the F1 of 0.538 is reported in \cite{adr}\footnote{The ADR task has been extensively explored in supervised settings \cite{adr-sup-1, adr-sup-2, smm4h}. However, the studies on semi-supervised ADR are limited \cite{adr-semisup}}.

Finally, often in the real world situations, practitioners who try to tackle a classification problem, may have a small training set for the task and a larger diverse training set in the similar domains. We tried to evaluate our model in such a scenario. Thus, we assumed \FLUDT dataset was the small training set which was available to do influenza surveillance in social media, and \PHMDT dataset was the bigger diverse dataset which was available for similar domains. In Table \ref{tbl-result-domain-adapt}, we report the results of domain adaptation in \FLUDT dataset, when we use \PHMDT dataset as the out-of-domain training data. We randomly sampled 500 positive and 500 negative examples from \PHMDT dataset and fine-tuned the models; then further fine-tuned them using the training folds of \FLUDT dataset, and finally used for labeling the \FLUDT test folds--we used this approach to prevent from the catastrophic forgetting phenomenon in neural networks \cite{catas}. The results signify that even with a moderately large balanced training set, a supervised model cannot outperform \METHOD\!.

\begin{table}
\begin{tabular}{|p{0.8in} |p{0.5in} |p{0.5in} |p{0.5in} |}\hline
\textbf{Model} & 
{\footnotesize \textbf{F1}} & 
{\footnotesize \textbf{Precision}} & 
{\footnotesize \textbf{Recall}} \\ \hline
\textit{BERT-TW} & 0.803 & 0.782 & 0.836 \\ \hline
\textit{\METHOD} & 0.810 & 0.813 & 0.813 \\ \hline
\end{tabular}
\caption{Domain adaptation results in \FLUDT dataset. 1000 training examples from \PHMDT dataset were randomly sampled--500 positives and 500 negatives--as the out-of-domain data.
} \label{tbl-result-domain-adapt}
\end{table}

In this study we defined problem decomposition, and showed that it has at least three important real-world applications in social media. Our model is defined for the tasks that are centered around a set of entities or concepts. \METHOD can be also regarded as an approach to incorporate domain knowledge into the machine learning models. In Section \ref{subsec:method-def}, we presented three arguments that explain why our model is effective: 1) The vector representation of words is smaller than the vector representation of documents. Thus, classification is easier over the words. 2) There are limited ways of using a word in a context. 3) Equipping the model with domain knowledge. The last argument, is based on the fact that we use domain understanding to impose a new inductive bias on the learner, through removing less important word features and targeting the pivotal entities in the task.

\section{Conclusions and Future Work} \label{sec:conclusion}

We proposed a novel semi-supervised model for classification tasks that are centered around specific entities or concepts. Our model is based on: (1) decomposing the problem into a set of sub-tasks, and (2) combining the results in a co-training framework. By leveraging domain knowledge to decompose problems, and employing co-training framework to reinforce the underlying classifiers, our model \METHOD is able to generalize well and outperform state-of-the-art classifiers in semi-supervised settings. We showed that our model is applicable to at least three important personal event detection problems, namely, Personal Health Mention detection, Crisis Report detection, and Adverse Drug Reaction monitoring. We also carried out extensive experiments and reported the performance of the model in various settings. The results indicate that \METHOD is able to consistently and significantly outperform state-of-the-art classifiers in the three mentioned tasks.


Our current research introduces three potential future work directions. First, investigating other tasks which may be decomposable. As we discussed in Section \ref{subsec:method-identify}, the tasks that are centered around entities and concepts can be potential targets. For instance, our model can be applied to the customer satisfaction task--where the mentions of human and the product can serve as candidate Key Concept Sets. The next two future directions are on the theory aspect of our method. One direction is to investigate the extent in which the choice of Key Concept Sets can impact the model performance. This will help us to understand whether our model can be applied to the tasks that the domain understanding is incomplete. Even though our experiments with a weak human mention detector showed promising results, we believe further investigation is required to understand if noisy Key Concept Sets can still be beneficial. And finally, the last future direction is to investigate the ways of automatically discovering Key Concept Sets.

\section*{Acknowledgments}
This work was funded by Emory University; also partially by NIH grant LM013014-02, NSF award IIS-\#1838200, and Google Cloud Platform research credits.




\bibliographystyle{ACM-Reference-Format}
\bibliography{sample-base}






\end{document}